\documentclass[a4paper,11pt]{article}

\usepackage{graphicx}  %%% for including graphics
\usepackage{url}       %%% for including URLs
\usepackage{times}
\usepackage[square,numbers]{natbib}
\usepackage[margin=25mm]{geometry}
\usepackage{amsmath}

\newcommand{\eg}{\emph{e.g.}, }       % for example
\newcommand{\ie}{\emph{i.e.}, }      % that is
\usepackage{booktabs}       % professional-quality tables
\usepackage{multirow}       % allowing multiple rows centered text in tables
\usepackage{xparse}         % diagonal text in tables
\NewDocumentCommand{\rot}{O{45} O{1em} m}{\makebox[#2][l]{\rotatebox{#1}{#3}}}

\title{Building Graph Representations of Deep Vector Embeddings}
\date{}

\author{Dario Garcia-Gasulla\\
       Barcelona Supercomputing Center (BSC)\\
       \texttt{dario.garcia@bsc.es}
  \and Armand Vilalta\\
       Barcelona Supercomputing Center (BSC)\\
       \texttt{armand.vilalta@bsc.es}
 \and Ferran Par\'{e}s\\
       Barcelona Supercomputing Center (BSC)\\
       \texttt{ferran.pares@bsc.es}
 \and Jonatan Moreno\\
       Barcelona Supercomputing Center (BSC)\\
       \texttt{jonatan.moreno@bsc.es}
 \and Eduard Ayguad\'{e}\\
  Barcelona Supercomputing Center (BSC)\\
Universitat Polit\`{e}cnica de Catalunya\\
       \texttt{eduard.ayguade@bsc.es}
 \and Jes\'{u}s Labarta\\
   Barcelona Supercomputing Center (BSC)\\
      Universitat Polit\`{e}cnica de Catalunya\\
       \texttt{jesus.labarta@bsc.es}
 \and Ulises Cort\'{e}s\\
         Barcelona Supercomputing Center (BSC)\\
Universitat Polit\`{e}cnica de Catalunya\\
       \texttt{ia@cs.upc.edu}
\and Toyotaro Suzumura\\
  Barcelona Supercomputing Center (BSC)\\
       IBM T.J. Watson\\
       \texttt{tsuzumura@us.ibm.com}
}

\begin{document}
\maketitle
\thispagestyle{empty}
\pagestyle{empty}

\begin{abstract}
    Patterns stored within pre-trained deep neural networks compose large and powerful descriptive languages that can be used for many different purposes. Typically, deep network representations are implemented within vector embedding spaces, which enables the use of traditional machine learning algorithms on top of them. In this short paper we propose the construction of a graph embedding space instead, introducing a methodology to transform the knowledge coded within a deep convolutional network into a topological space (\ie a network). We outline how such graph can hold data instances, data features, relations between instances and features, and relations among features. Finally, we introduce some preliminary experiments to illustrate how the resultant graph embedding space can be exploited through graph analytics algorithms.
\end{abstract}

\section{Introduction}

Deep learning models build large and rich data representations by finding complex patterns within large and high-dimensional datasets. At the end of a deep learning training procedure, the learnt model can be understood as a data representation language, where the pattern learnt by each neuron within the deep model represents a word of such language. Extracting and reusing the patterns learnt by a deep neural network (DNN) is a subfield of deep learning known as transfer learning. Transfer learning from a pre-trained DNN can be used to initialize the training of a second DNN from a non-random state, improving performance over randomly initialized networks \cite{xu2015augmenting,branson2014bird,liu2016deepfood}, and also enabling the training of DNNs for domains with limited amount of data \cite{ge2017borrowing,simon2015neural}. These two settings, where the purpose of the transfer learning process is to train a second DNN, are cases of transfer learning for fine tuning. A different purpose of transfer learning is to extract deep representations so that alternative machine learning methods can be run on top of those. This is commonly known as \textit{transfer learning for feature extraction}, and is the main topic of this paper.

%The DL training process optimizes the model parameters (\ie the neuron weights) with the goal of minimizing the error on a complex task (\eg image classification).
%, where the complexity of each term in the language is related with the location of the corresponding neuron within the network (typically, neurons in deep layers correspond to more complex patterns or terms).

Extracted DNN representations are typically implemented through vectors, where the length of the vector equals to the number of neural features being used. These vector embedding spaces have been used to feed classifiers based on the instance-attribute paradigm (\eg Support Vector Machines) \cite{azizpour2016factors,garcia2017out}. Instead, in this paper we propose a \textit{graph} based representation of those same embeddings spaces, with the goal of running a different family of algorithms; those based on the instance-instance paradigm, such as community detection algorithms. Graph or network based algorithms focus on the associations among instances to find topologically coherent patterns. These are significantly different from the patterns that can be found using algorithms focused on the associations among instances and attributes. This paper describes a methodology for building a graph representation of neural network embeddings (in \S\ref{sec:graph}), and reports the performance of a community detection algorithm processing the resultant graph (in \S\ref{sec:exp}).

\section{Graph Representation of Vector Embeddings}\label{sec:graph}

Vector embeddings are often built by capturing the output of a single layer close to the output of a deep convolutional neural network (CNN), typically a fully-connected layer \cite{sharif2014cnn,gong2014multi,donahue2014decaf,mousavian2015deep,ren2017generic}. However, earlier layers can also be useful for characterizing data, particularly when this data is not strongly related with the data that was used for pre-training the model \cite{azizpour2016factors}. Another difference between later layers and the rest in a typical CNN trained for classification is that layers close to the output exhibit a strongly discriminative behavior, activating sporadically and in a crisp manner (either strongly activating or not at all). In contrast, neurons of layers further from the output show a more descriptive behavior, activating frequently and in a fuzzy way \cite{garcia2017behavior}.

In this paper we generate a graph representation which captures data properties in a topological space (\ie through vertices and their associations). Using a single layer embedding for that purpose would result in a rather poor topology, as fully-connected layers represent but a portion of all patterns learnt by the DNN, and only a small subset of neurons activate for each data instance. To guarantee that the graph representation contains a topology rich enough as to empower network analysis algorithms we use the full-network embedding \cite{garcia2017out}, which produces a vector embedding including all convolutional and fully-connected layers of a CNN. This results in a much larger embedding space (composed by tens of thousands of dimensions), allowing us to generate larger and richer graphs.

Next we briefly introduce the full-network embedding, and at the end of this section we describe how we generate a graph to capture the embedding space within its topology.

\subsection{Full-network Embedding}

\begin{figure}[t]
  \centering
  \includegraphics[width=\linewidth]{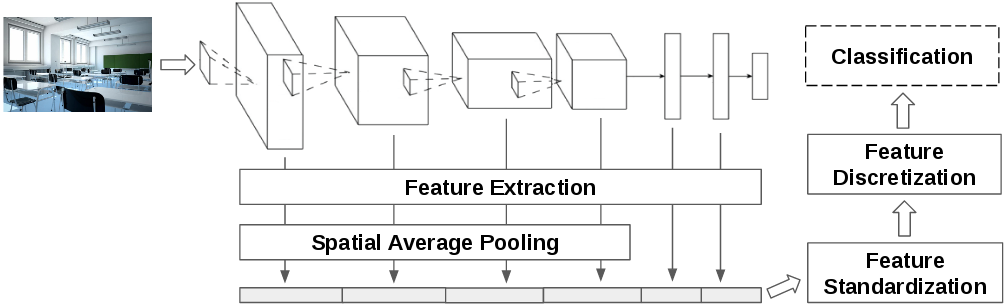}
  \caption{Overview of the Full-network embedding generation workflow.}
  \label{fig:workflow}
\end{figure}

The full-network embedding (FNE) generates a representation of an input data instance by capturing the activations it produces at every convolutional and fully-connected layer within a CNN. In order to integrate the features found at layers of different depth, size and nature, the FNE includes a set of processing steps. An overview of these can be seen in Figure \ref{fig:workflow}. 

After the extraction of activations, the FNE applies a spatial average pooling on the activations coming from convolutional layers. As a result, neurons of convolutional layers will generate a single value in the embedding (as neurons from fully-connected layers do). After the spatial pooling, the FNE includes a feature standardization step. The goal of this transformation is to normalize the values of the different neurons, so that each neuron has a coherent range of activations regardless of its type or location within the CNN. Otherwise, the activations from the layers close to the output would dominate the representation, as neurons from these layers typically activate with much more strength.

Finally, the FNE discretizes features, mapping all values to either -1, 0 or 1. This process reduces noise and regularizes the embedding space. In the FNE, this discretization is done with a pair of constant thresholds $(-0.25, 0.15)$ which determine if a feature is relevant by presence (1 implies an abnormally high activation) or is relevant by absence (-1 implies an abnormally low activation) for a given input data instance \cite{garcia2017out}. In our experiments we set more demanding thresholds $(-2.0, 2.0)$, to make sure that the degree of sparsity of the graph is appropriate for network analysis methods.

\subsection{Graph Representation}

The FNE generates a vector representation of each data input, where feature values are either -1, 0 or 1. Those values represent the relevance of a specific neural filter for a given data input, indicating feature relevance by absence, feature irrelevance, and feature relevance by presence, respectively. In this paper we consider building a topology based representation (\ie a graph) of such embedding, and processing it through algorithms exploiting instance-instance relations (\eg community detection) to exploit the information encoded topologically.

\subsubsection{Vertices}

Let us start defining what composes the vertices (V) of our graph representation G. From the discrete three-valued full-network embedding we extract both data instances (\eg images of a dataset) and model features (\ie neural filters of the CNN). Each data instance is represented in the graph $G$ as a unique vertex (of type image vertex or $V_i$). Features however, may be relevant by presence or by absence. We choose to capture this dichotomy by creating two vertices in the graph for each feature; one of those vertices will represent the relevant activation of the feature (positive feature vertex or $V^+_f$), while the other will represent the relevant lack of activation of the feature (negative feature vertex or $V^-_f$). As a result, the number of image vertices in the graph ($|V_i|$) will be equal to the number of input data images, while the number of positive and negative feature vertices ($|V^+_f|$ and $|V^-_f|$ separately) will be equal to the number of features in the original vector embedding.

\subsubsection{Image-Feature Edges}
Let us now define the edges (E) of our graph representation G. The implementation of edges between image vertices and feature vertices ($E_{if}$) is straight-forward. We add an edge between a vertex image $v_i$ and a positive feature vertex $v^+_f$ when the corresponding value in the full-network embedding is 1. Analogously, we add an edge between a vertex image $v_i$ and a negative feature vertex $v^-_f$ when the corresponding value in the full-network embedding is -1. Edges between image vertices and positive feature vertices indicate that the occurrence of the feature is relevant for describing the image, while edges between image vertices and negative feature vertices indicate that the no occurrence of the feature is relevant for describing the image. Feature values of 0 identify irrelevancy, which is why these are not coded into the graph.

% TODO: integrar amb el text
%The complete set of edges can be divided in feature-image and feature feature edges ($E=\{E_{if},E_{ff}\}$). 

\subsubsection{Feature-Feature Edges}

One of the benefits of using a graph-based representation for the embedding is that, unlike vector representations, it allows us to extend the embedding space to include feature-feature relations. In the graph this corresponds to the creation of edges between feature vertices ($E_{ff}$). To identify relations between features we consider the weights ($\mathcal{W}$) of the pre-trained CNN. If the weight associating a neuron $f_{l+1}$ with a neuron from the previous layer $f_l$ ($\mathcal{W}(f_{l+1},f_{l})$) is abnormally high in the context of all weights of $f_{l+1}$ with the previous layer ($\mathcal{W}(f_{l+1}, F_{l})$), there exists a positive correlation between both neurons. Similarly, if $\mathcal{W}(f_{l+1},f_{l})$ is abnormally low in the context of $\mathcal{W}(f_{l+1}, F_{l})$, there exists a negative correlation between neurons $f_{l+1}$ and $f_{l}$. To identify what counts as abnormally high or abnormally low, we compute the mean ($\mu$) and the standard deviation ($\sigma$) of the weights that associate a feature $f_{l+1}$ from a given layer with every feature from the previous layer $f_i \in F_l$ (in the case of convolutional filters we previously sum all the weights of the receptive field). If the weight associating $f_{l+1}$ with one of the features of the previous layer is above that mean a given number of standard deviations, then such relation is abnormally high (see Equation \ref{eq:rel}). An abnormally low relation is defined analogously.

We implement positive correlations in the graph by adding an edge between the positive feature vertex of $f_{l}$ and the positive feature vertex of $f_{l+1}$ ($e_{ff}(v^+_{f_l},v^+_{f_{l+1}})$), and another edge between the negative feature vertex of $f_{l}$ and the negative feature vertex of $f_{l+1}$ ($e_{ff}(v^-_{f_l},v^-_{f_{l+1}})$). Negative correlations are implemented by adding an edge between the positive feature vertex of $f_{l}$ and the negative feature vertex of $f_{l+1}$ ($e_{ff}(v^+_{f_l},v^-_{f_{l+1}})$), and another edge between the negative feature vertex of $f_{l}$ and the positive feature vertex of $f_{l+1}$ ($e_{ff}(v^-_{f_l},v^+_{f_{l+1}})$).

%($\{e_{ff}(v^+_{f_l},v^+_{f_{l+1}}), e_{ff}(v^-_{f_l},v^-_{f_{l+1}})\}\in E_{ff}$). However, if such relation is negative (the weights between $f_{l}$ and $f_{l+1}$ are abnormally low), we will add an edge between the positive feature vertex of $f_{l}$ and the negative feature vertex of $f_{l+1}$, and also between the negative feature vertex of $f_{l}$ and the positive feature vertex of $f_{l+1}$ ($\{e_{ff}(v^+_{f_l},v^-_{f_{l+1}}),e_{ff}(v^-_{f_l},v^+_{f_{l+1}})\}\in E_{ff}$). 
%For simplicity we notate the two edges defined due to a positive relation between features $f_i$ and $f_j$ as $e^+_{ff}(f_i,f_j)$ and analogously for the ones due to a negative relation $e^-_{ff}(f_i,f_j)$.

%\begin{equation}\label{eq:high}
%weight(f_{l+1},f_{l})>\mu(f_{l+1})+\sigma(f_{l+1})*k
%\end{equation}

%\begin{equation}\label{eq:low}
%weight(f_{l+1},f_{l})<\mu(f_{l+1})-\sigma(f_{l+1})*k
%\end{equation}

\begin{equation}\label{eq:rel}
 \begin{array}{crr}
        %\text{High relation:} &
        %e^+_{ff}(f_{l+1},f_{l}) \in E^+_{ff}
        \{e_{ff}(v^+_{f_l},v^+_{f_{l+1}}), e_{ff}(v^-_{f_l},v^-_{f_{l+1}})\}\in E_{ff}
        & \text{iff}  & \mathcal{W}(f_{l+1},f_{l})>\mu(\mathcal{W}(f_{l+1}, F_{l}))+\sigma(\mathcal{W}(f_{l+1}, F_{l}))*k\\
        %\text{Low relation:} & 
        %e^-_{ff}(f_{l+1},f_{l}) \in E^-_{ff}
        \{e_{ff}(v^+_{f_l},v^-_{f_{l+1}}),e_{ff}(v^-_{f_l},v^+_{f_{l+1}})\}\in E_{ff}
        & \text{iff} & \mathcal{W}(f_{l+1},f_{l})<\mu(\mathcal{W}(f_{l+1}, F_{l}))-\sigma(\mathcal{W}(f_{l+1}, F_{l}))*k\\
 \end{array}
\end{equation}
%TODO: feature-feature threshold? definició o valor o algo

The parameter $k$ of Equation\ref{eq:rel} regulates the sparsity of feature-feature edges. A higher $k$ value will only accept edges between very strongly connected pairs of neurons, as the associated weights must be a larger number of standard deviations above the mean. In all our experiments we set $k$ to $1.5$.

At this point we can finally define our graph representation as $G= (V,E)$ where $E=E_{if}\cup E_{ff}$ and $V=V_i\cup V^+_f\cup V^-_f$.

\section{Experiments}\label{sec:exp}

To evaluate our graph representation of the deep embedding space we use the VGG16 CNN architecture \cite{simonyan2014very}, pre-trained on the ImageNet \cite{russakovsky2015imagenet} dataset. We process four different datasets through this pre-trained model. Details on the datasets are shown in Table \ref{tab:datasets}:

\begin{itemize}
\item The MIT Indoor Scene Recognition dataset \cite{quattoni2009recognizing} (\textit{mit67}) consists of different indoor scenes to be classified in 67 categories.
\item The Oxford Flower dataset \cite{nilsback2008automated} (\textit{flowers102}) is a fine-grained dataset consisting of 102 flower categories.
\item The Describable Textures Dataset \cite{cimpoi2014describing} (\textit{textures}) is a database of textures categorized according to a list of 47 terms inspired from human perception.
\item The Oulu Knots dataset \cite{silven2003wood} (\textit{wood}) contains knot images from spruce wood, classified according to Nordic Standards.
\end{itemize}

For each of the images of those datasets we obtain the full-network embedding. In the case of the VGG16 architecture, the embedding generates vectors of 12,416 features. Based on those, we build the graph representation, as previously described.

\begin{table}[t]
    \caption{Properties of all datasets computed}% as used in our experiments}
    \label{tab:datasets}
    \centering
    \begin{tabular}{crrr}
        \toprule
        Dataset    & \#Images  & \#Classes    &   \#Images per class \\
        \midrule
        \textit{mit67}       &   6,700   &   67  &   100     \\
        \textit{flowers102}  &   8,189   &   102 &   40 - 258\\
        \textit{textures}    &   5,640   &   47  &   120     \\
        \textit{wood}        &   438     &   7   &   14 - 179  \\
        \bottomrule
    \end{tabular}
\end{table}

We explore the graph-representation by running a community detection algorithm on top of it. Particularly, we use the Fluid Communities (FluidC) algorithm \cite{pares2017fluid}. We chose this algorithm because its based on the efficient label propagation methodology while outperforming the traditional LPA algorithm, because it allows us to specify the number of clusters we wish to find, and because it can be easily adapted to the specific needs of our experiments. These specific needs regard mostly cluster initialization and maintainance. Since the graph is composed by both images and features vertices, but only images have an associated label, clusters should be evaluated considering only the image vertices found in a final community. For this same reason, we must ensure that all communities found contain at least one image vertex. This is done by modifying the FluidC algorithm, forcing it to initialize communities on image vertices, and by making sure that a community contains at least one image vertex at all times. 

We measure the quality of the found clusters by measuring the similarity between the found communities and the original dataset labels. We use both the normalized mutual information measure (NMI) and the adjusted mutual information (AMI). The properties of the graph generated for each dataset and the performance results are shown in Table \ref{tab:graphs}.

All experiments were done using the VGG16 model pre-trained on ImageNet2012, freely available online. The feature extraction process was done with Caffe. The execution of the graph algorithm was done using NetworkX v2.0, which includes FluidC.

\begin{table}[t]
    \caption{Properties of the graphs built from the deep embedding spaces, and quality of the communities found by the FluidC algorithm measured in NMI and AMI.}
    \label{tab:graphs}
    \centering
    \begin{tabular}{ccccc}
        \addlinespace[10pt]
        \toprule
        Dataset & \textit{mit67} & \textit{flowers102} & \textit{textures} & \textit{wood}\\
        \midrule
        $|V|$  & 30,186 &  33,010  & 30,459    & 25,259  \\
        $|E|$  & 8,396,939 & 9,654,464  &  8,561,314    & 5,610,832  \\
        \midrule
        NMI  & 0.44 & 0.54 & 0.42 & 0.26 \\
        AMI  & 0.36 & 0.46 & 0.37 & 0.20 \\
        \bottomrule
    \end{tabular}
\end{table}

\section{Related Work}\label{sec:sota}

The relation between graphs and deep neural networks have been previously explored, but most contributions do so from a different perspective. While our proposal is to obtain a graph representation of the embedding produced by a CNN when processing an image, most related work focuses on training DNNs for processing graph data. For example, DeepWalk \cite{perozzi2014deepwalk} uses random walks in a graph (\eg Flickr or YouTube networks) to feed a SkipGram model, and then evaluate community detection methods on those graphs. Similarly to DeepWalk, the work of \citet{cao2016deep} also processes graphs as input, but it uses a probabilistic approach on weighted graphs to feed an autoencoder. In contrast, we are performing community detection on a dataset of images, something that, to the best of our knowledge, had not been done before.

\section{Conclusions}\label{sec:conc}

The presented methodology is a first step towards building graph-based representations of deep CNN embeddings. We detail how to include in such a graph both images and features, and how to topologically codify image-feature and feature-feature relations. By doing so we make deep knowledge available to a large set of learning algorithms (network analysis tools) which may exploit those representations in a completely different manner.

The results reported are encouraging, as a topology based algorithm such as FluidC is capable of identifying relevant communities of images using only topological information. The clusters that can be found through network analysis tools are significantly different than the clusters that can be found through more traditional algorithms (\eg Kmeans) running on a vector representation. Indeed, while algorithms like FluidC exploit the paths among vertices, algorithms like Kmeans are based on distance measures which can hardly integrate the information of all possible paths in a graph. The possibility of running a different type of algorithms makes this novel approach interesting, as it opens the door at exploiting and reusing the knowledge coded within deep pre-trained neural models in a completely new way. This is but a small step towards a better understanding of deep representations, and how to exploit all the knowledge these encode for a wider variety of purposes.

\section{Future Work}

The experiments shown in this paper are highly exploratory, with the purpose of validating the feasibility of the hypothesis (\ie that deep neural representations can be implemented as a graph coherently). Analytically, it would be desirable to compare the performance of the proposed approach with alternatives, such as applying a clustering algorithm on the original vector embeddings. However, clustering evaluation is a controversial topic, as there is not a single solution which may be considered to be the universal ground truth.

A more relevant evaluation of the approach should be based on an analysis of the semantics captured by the graph. For that purpose we are considering the extension of the model to include directed edges, weighted edges, and ontological relations. By doing so one could execute inference methods on top of the representation, and measure how rich and usable the semantics captured in the graph actually are.

\section*{Acknowledgements}
This work is partially supported by the Joint Study Agreement no. W156463 under the IBM/BSC Deep Learning Center agreement, by the Spanish Government through Programa Severo Ochoa (SEV-2015-0493), by the Spanish Ministry of Science and Technology through TIN2015-65316-P project, by the Generalitat de Catalunya (contracts 2014-SGR-1051), and by the Core Research for Evolutional Science and Technology (CREST) program of Japan Science and Technology Agency (JST).

\bibliographystyle{chicago}
\bibliography{graph_rep}

\end{document}